\documentclass{article}
\usepackage{times}
\usepackage{graphicx} 
\usepackage{subfigure}
\usepackage{natbib}
\usepackage{algorithm}
\usepackage{algorithmic}
\usepackage{hyperref}

\usepackage[accepted]{icml2017}
\usepackage{url}
\usepackage{latexsym}
\usepackage{tabularx}
\usepackage{multirow}
\usepackage{amsmath}
\usepackage[normalem]{ulem}
\usepackage{epsfig}
\usepackage{amsmath}
\usepackage{amsfonts}
\usepackage{amsthm}
\usepackage{amssymb}

\usepackage[group-separator={,},group-minimum-digits={3}]{siunitx}

\newcommand{\xhat}{\widehat{x}}

\newcommand{\setB}{\mathcal{B}}
\newcommand{\E}{\text{E}}
\newcommand{\Var}{\text{Var}}

\newcommand{\kk}{{(k)}}
\newcommand{\vx}{\mathrm{x}}

\newcommand{\vb}{\mathrm{b}}

\icmltitlerunning{Minimal Effort Back Propagation for Convolutional Neural Networks}

\begin{document}

\twocolumn[
\icmltitle{Minimal Effort Back Propagation for Convolutional Neural Networks}


\begin{icmlauthorlist}
\icmlauthor{Bingzhen Wei}{eecs,moe}
\icmlauthor{Xu Sun}{eecs,moe}
\icmlauthor{Xuancheng Ren}{eecs,moe}
\icmlauthor{Jingjing Xu}{eecs,moe}
\end{icmlauthorlist}

\icmlaffiliation{eecs}{School of Electronics Engineering and Computer Science, Peking University, China}
\icmlaffiliation{moe}{MOE Key Laboratory of Computational Linguistics, Peking University, China}


\icmlkeywords{acceleration, sparse learning, neural networks, CNN, overfitting}

\vskip 0.3in
]

\printAffiliationsAndNotice{}

\begin{abstract}
As traditional neural network consumes a significant amount of computing resources during back propagation,
\citet{Sun2017mePropSB} propose a simple yet effective technique to alleviate this problem. In this technique, only a
small subset of the full gradients are computed to update the model parameters. In this paper we extend this
technique into the Convolutional Neural Network(CNN) to reduce calculation in back propagation, and the surprising
results verify its validity in CNN: only 5\% of the gradients are passed back but the model still achieves the
same effect as the traditional CNN, or even better. We also show that the top-$k$ selection of gradients
leads to a sparse calculation in back propagation, which may bring significant computational benefits for high computational
complexity of convolution operation in CNN.
\end{abstract}

\section{Introduction}
Convolutional Neural Networks (CNN) have achieved great success in many fields (especially in visual recognition tasks),
such as object classification \cite{krizhevsky2012imagenet}, face recognition \cite{taigman2014deepface}. Although the special
network architecture of CNN makes it possible to get abstract features layer by layer, high computational complexity of
the convolution computation consumes a large amount of the computing resources, and this problem turns CNN into a compute-intensive model.
CNN needs to do convolution computation via a matrix multiplication operation for each convolutional layer in the forward
propagation, on the other hand, almost the same amount of computation is needed during back propagation. Because of this, a method
that could reduce calculation will be of great help for reducing the time consumption in both the training process and the inference process.

\citet{Sun2017mePropSB} propose a \emph{minimal effort} back propagation method to reduce calculation in back propagation, which 
called \emph{meProp}. The idea is to compute only a very small but critical portion of the gradient information, and update
only the corresponding minimal portion of the parameters in each learning step. In this way, we only update the highly relevant parameters,
while others stay untouched. Hence, this technique results in sparse gradients and sparse update. In other words, fewer gradients are passed
back and only $k$ rows or columns (depending on the layout) of the weight matrix are modified. The experiments also show that models using
\emph{meProp} are more robust and less likely to be overfitting.

We extend this technique to Convolutional Neural Network, which we call \emph{meProp}-CNN, to reduce calculation in back propagation of CNN.
In back propagation of CNN, the convolution operation is transformed into matrix multiplication operations as in forward propagation.
As in most neural networks, the matrix multiplication operation consumes more computing resources than other operations, such as plus,
minus, and so on. To address this issue, we apply \emph{meProp}-CNN in CNN, just like the \emph{meProp} in feedforward NN model (MLP)
in \citet{Sun2017mePropSB}.

The differences from \emph{meProp} in MLP and the contributions of this work are as follows:

\begin{figure*}[t]
\begin{center}
\begin{tabular}{c}
    \includegraphics[width=0.8\hsize]{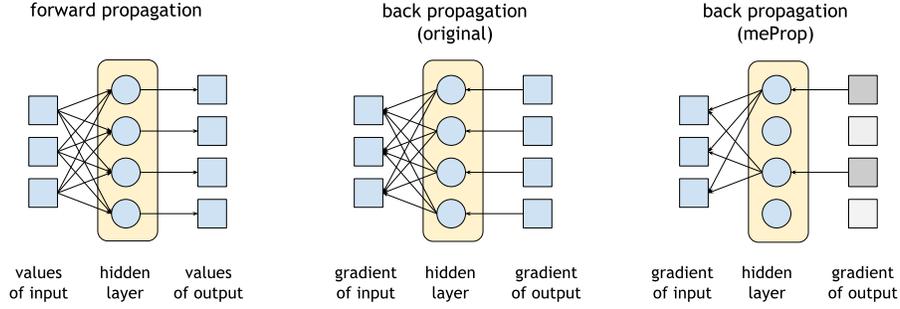}
\end{tabular}
\caption{An illustration of meProp.}\label{fig1}
\end{center}
\vspace{-0.15in}
\end{figure*}

\begin{itemize}
    \item
    Compared with the linear transformation in MLP, the convolution operation in CNN is a unique operation, and this characteristic leads to a
    different behavior during parameters updation. This will be explained in detail in Section~\ref{section:meProp-CNN}.
    \item
    We implement a new sparse back propagation method for CNN to reduce calculation, which makes the complex convolution computation
    transformed into sparse matrix multiplication. In this way, the proposed method can outperform the original methods.
    \item
    We enhance \emph{meProp} technique with momentum method to get more stable results. This is an optional method in our experiments.
\end{itemize}

\section{Method}
We introduce \emph{meProp} technique into the Convolutional Neural Network to reduce calculation in back propagation. The
forward process is computed as usual, while only a small subset of gradients are used to update the parameters. Specifically,
we select top-$k$ elements to update parameters and the rest are set to $0$, which is similar to the \emph{Dropout}~\cite{srivastava2014dropout} technique.
And we do find that model with \emph{meProp} is more robust and less likely to be overfitting. We first present the proposed
method and then describe the implementation details.

\begin{figure*}[t]
\begin{center}
\begin{tabular}{c}

    \includegraphics[width=0.3\hsize]{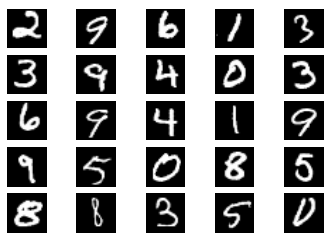}\\

\end{tabular}
\caption{MNIST sample images}\label{fig_data}
\end{center}
\vspace{-0.05in}
\end{figure*}

\subsection{meProp}
\label{section:meProp}
We first introduce \emph{meProp} in feedforward neural network. For simplicity, a linear transformation unit
is brief enough for us to explain and understand the detail of the proposed method:
\begin{equation}\label{linear}
y= W  x
\end{equation}
\begin{equation}\label{nonlinear}
   z=\sigma(y)
\end{equation}
where $W \in R^{n \times m},  x \in R^{m},  y \in R^{n},  z \in R^{n}$, $m$ is the dimension of the input vector, $n$ is the dimension of the output vector, and $\sigma$ is a non-linear function (e.g., \emph{relu}, \emph{tanh}, and \emph{sigmoid}). During back propagation, we need to compute the gradient of the parameter matrix $W$ and the input vector $x$:
\begin{equation}\label{der1}
   \frac{\partial  z}{\partial  W_{ij}}=\sigma^{'}_{i}{ x}^{T}_{j} \ \ \ (1\leq i \leq n, 1\leq j \leq m)
\end{equation}
\begin{equation}\label{der2}
   \frac{\partial  z}{\partial  x_{i}}=\sum_{j}{{ W}^{T}_{ij}\sigma^{'}_{j}} \ \ \ (1\leq j\leq n, 1\leq i \leq m)
\end{equation}
where $\sigma^{'}_i \in R^{n}$ means $\frac{\partial  z_i}{\partial  y_i}$. We can see that the computational cost of back propagation is directly proportional to the dimension of output vector $n$.

The proposed meProp uses approximate gradients by keeping only top-$k$ elements based on the \emph{magnitude values}. That is, only the top-$k$ elements with the largest absolute values are kept. For example, suppose a vector $v=\langle 1,2,3,-4 \rangle$, then $top_2(v)= \langle 0,0,3,-4 \rangle$.
We denote the indices of vector $\sigma^{'}(y)$'s top-$k$ values as $\{t_1, t_2, ..., t_k\}(1 \leq k \leq n)$, and the approximate gradient of the parameter matrix $W$ and input vector $x$ is:
\begin{equation}\label{der3}
   \frac{\partial z}{\partial W_{ij}} \leftarrow \sigma^{'}_{i}x^{T}_{j} \ \  \text{ if } \ \ i \in \{t_1, t_2, ..., t_k\} \ \ \text{ else } \ \ 0
\end{equation}
\begin{equation}\label{der4}
   \frac{\partial z}{\partial x_{i}} \leftarrow \sum_{j}{W^{T}_{ij}\sigma^{'}_{j}} \ \  \text{ if } \ \ j \in \{t_1, t_2, ..., t_k\} \ \ \text{ else } \ \ 0
\end{equation}
As a result, only $k$ rows or columns (depending on the layout) of the weight matrix are modified, leading to a linear reduction ($k$ divided by the vector dimension) in the computational cost.

Figure~\ref{fig1} is an illustration of meProp for a single computation unit of neural models. The original back propagation uses the full gradient of the output vectors to compute the gradient of the parameters. The proposed method selects the top-$k$ values of the gradient of the output vector, and backpropagates the loss through the corresponding subset of the total model parameters.

As for a complete neural network framework with a loss $L$, the original back propagation computes the gradient of the parameter matrix $W$ as:
\begin{equation}\label{der5}
    \frac{\partial L}{\partial W} = \frac{\partial L}{\partial y} \cdot \frac{\partial y}{\partial W}
\end{equation}
while the gradient of the input vector $x$ is:
\begin{equation}\label{der6}
    \frac{\partial L}{\partial x} = \frac{\partial y}{\partial x} \cdot \frac{\partial L}{\partial y}
\end{equation}
The proposed meProp selects top-$k$ elements of the gradient $\frac{\partial L}{\partial y}$ to approximate the original gradient, and passes them through the gradient computation graph according to the chain rule. Hence, the gradient of $W$ goes to:
\begin{equation}\label{der5}
    \frac{\partial L}{\partial W} \leftarrow  top_k(\frac{\partial L}{\partial y}) \cdot \frac{\partial y}{\partial W}
\end{equation}
while the gradient of the vector $x$ is:
\begin{equation}\label{der6}
    \frac{\partial L}{\partial x} \leftarrow  \frac{\partial y}{\partial x} \cdot top_k(\frac{\partial L}{\partial y})
\end{equation}

\begin{figure*}[h]
\begin{center}
\begin{tabular}{c}
    \includegraphics[width=1.0\hsize]{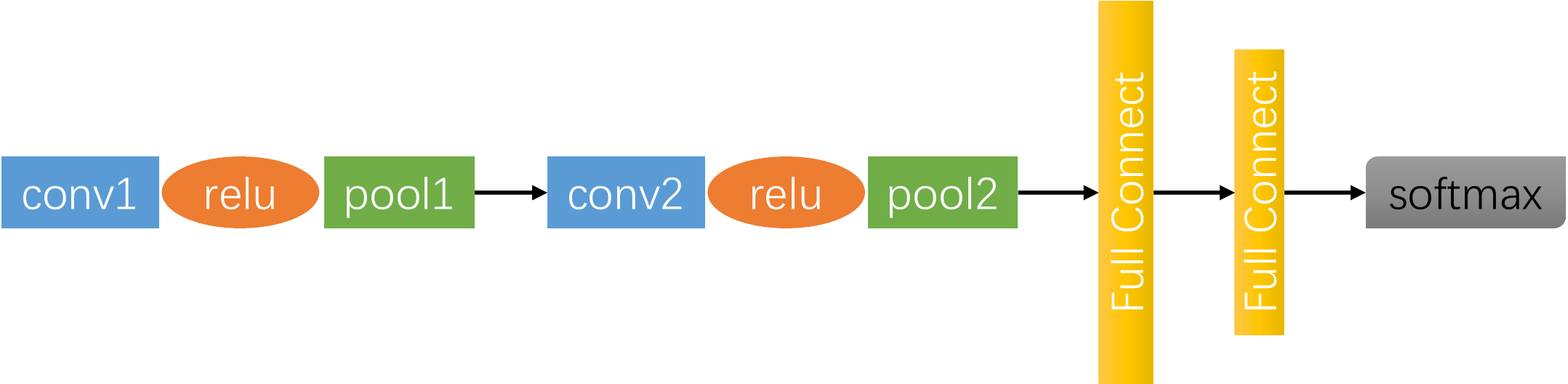}
\end{tabular}
\caption{CNN structure of MNIST task.}\label{fig_net}
\end{center}
\vspace{-0.15in}
\end{figure*}

\begin{figure}[h]
\begin{center}
\begin{tabular}{c}
    \includegraphics[width=1.0\hsize]{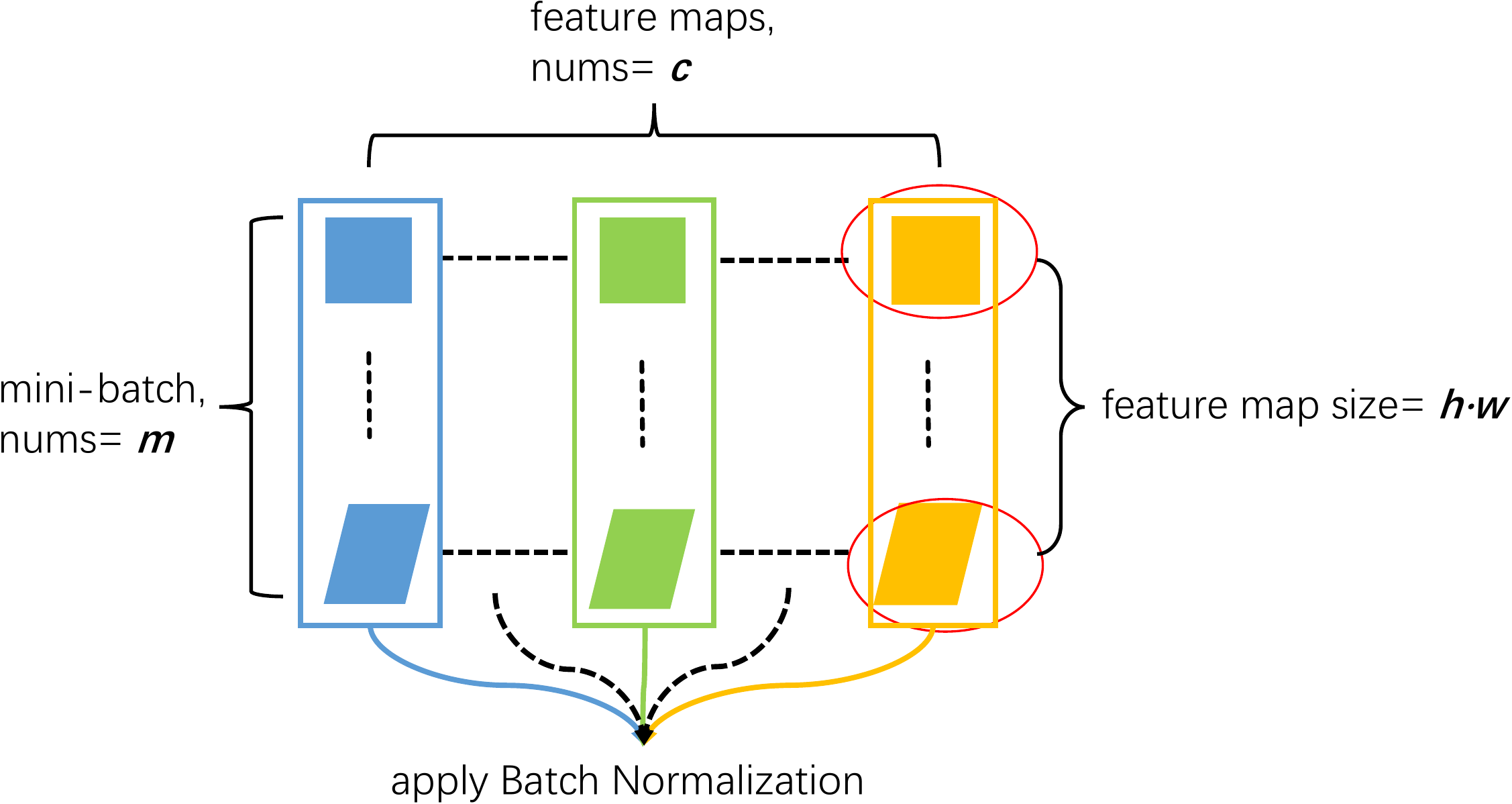}
\end{tabular}
\caption{Illustrate the output shape of convolutional layer. There are $m$ samples in one mini-batch, and $c$ filters will
        generate $c$ feature maps (the size of each is $h\cdot w$).}\label{fig_2}
\end{center}
\vspace{-0.15in}
\end{figure}

\subsection{meProp-CNN}
\label{section:meProp-CNN}
The forward propagation process is the same as it is in feed forward neural networks.
Compared with linear transformation in Section~\ref{section:meProp}, convolution computation in CNN is a unique operation, and this feature leads to a
different behavior during parameters update. In MLP, only the corresponding critical portion of the parameters are updated in
each learning step, which means only several rows or columns (depending on the layout) of the weight matrix are modified. But
it is not necessarily like this in \emph{meProp}-CNN. \emph{meProp} operation only generates sparse matrix in intermediate
gradients. Take a simple convolution computation for example:
\begin{equation}\label{convolution1}
y= W \otimes x
\end{equation}
We use $\otimes$ to denote the operation of a convolution for CNN, and $W$ to denote parameters of the filters while $x$ to denote the input of the current layer, which is the output of the previous layer.
To get the gradient of $W$, $\frac{\partial L}{\partial W}$, we need to compute $\frac{\partial L}{\partial y}$
as:
\begin{equation}\label{convolution2}
\frac{\partial L}{\partial W} = \frac{\partial L}{\partial y} \otimes x^{T}
\end{equation}
Then we apply \emph{meProp} to Eq.~\eqref{convolution2}:
\begin{equation}\label{convolution2-top}
\frac{\partial L}{\partial W} = top(\frac{\partial L}{\partial y}) \otimes x^{T}
\end{equation}
Here the operands are all transformed into right matrix shape as needed. 
Note that after convolution computation, the gradients of $W$ are probably not as sparse as in MLP model. This is determinated by the difference between convolution operation and linear transformation. So the benefit we get in \emph{meProp}-CNN is sparse matrix operation, and it is necessary to verify the validity of \emph{meProp} in CNN architecture.

Dense matrix operations in CNN  consume most of the time of back propagation. To address this we propose
\emph{meProp}-CNN technique, which will lead to sparse matrix operations, and we will benefit from this transformation.
We apply the proposed method in every convolution filter respectively.
As in \citet{Sun2017mePropSB}, for other element-wise operations (e.g., activation functions), the original back
propagation procedure is kept, because those operations are already fast enough compared with matrix-matrix or matrix-vector
multiplication operations. 

In this paper, we use top-$k$ as the ratio of gradients that is selected in one hidden layer. For example, if we set top-$k$ = 5\%,
then 50 gradients will be selected for a layer under the dimension of 1000.

As illustrated in Figure~\ref{fig_net}, this is a common architecture for Convolutional Neural Networks, and we apply our method in 
convolutional layers: $conv1$, $conv2$ in Figure~\ref{fig_net}, and full gradients are passed back in fully--connected layers. 
Note that in Eq.~\eqref{convolution1} the operations of convolution are transformed into matrix-matrix or matrix-vector multiplication operations, then we can rewrite Eq.~\eqref{convolution1} as: 
\begin{equation}\label{conv2}
    y= W^{'}x^{'}
\end{equation}
Since we conduct our experiments based on Tensorflow, we need not make these Matrix transformation manually. The operands are all 
transformed into right matrix shape properly by Tensorflow framework, all we need to do is to apply our method on the gradients.
Consider that chain rule is used in back propagation, we need to get the gradients of the parameter matrix $W^{'}$ and the input vector $x^{'}$.
Heretofore, the operations of convolution are transformed into matrix-matrix or matrix-vector multiplication operations so the process 
of back propagation is similar with \emph{meProp} decribed in Section~\ref{section:meProp}.\\ 
But always remember that convolution is different from other operations because of its weight sharing: all of the units in a feature map share 
the same parameters, namely, the same filters weights. Weight sharing mechanism makes CNN powerful to extract abstract features layer by layer. 
The filters slide over local patches of feature maps from current layer and then generate feature maps for the next layer. On the other hand,
different filters hold different parameters and work for the model independently, this urges us to obey the same property when we apply proposed
method during back propagation. That is to say, we should select top-k gradients for every feature map respectively, rather than mixing all feature
 maps together then choosing from them. Concretely, the output $y$ of the current convolutional layer contains $m$ feature maps which are generated 
 by $m$ filters respectively in forward propagation, then in back propagation we will get gradient matrix $\frac{\partial L}{\partial y}$, which has
  the same shape as matrix $y$. Slice the gradient matrix into $m$ pieces corresponding to the $m$ filters in forward propagation and then apply 
  top-$k$ selection respectively, that is the top-$k$ elements with the largest absolute values are kept. One thing that should be pointed out is that
   we do not apply top-$k$ selection on the current gradient matrix directly, instead we take into account the historical gradients scale. We achieve this
    by applying exponential decay, formally, the accumulated gradients matrix is updated as:\\
\begin{equation}\label{momentum}
    run\_grad = decay\times run\_grad + (1-decay)\times |\frac{\partial L}{\partial y}|
\end{equation}
Then we select top-$k$ elements based on the new matrix $run\_grad$ and the unselected elements are set to 0. The selected sparse matrix will replace the original gradient matrix to complete the rest work for previous layers, as Eq.~\eqref{der5} and Eq.~\eqref{der6} do.

Also note that we use \emph{relu} as the activation function, and one property of \emph{relu} is that it tends to lead to sparsity, more or less. Then we check the CNN to find out how much sparsity \emph{relu} and Max-pooling layer contribute to the gradients. When we train CNN for
1 iteration, the sparsity(the ratio of nonzero values) of 3 sparse layers(conv1,conv2 and the first fully--connected layer) is 23\%, 3\% and 50\%, respectively. In contrast, the sparsity of sigmoid is 25\%, 25\% and 99\%, and the rate 25\% is related to the kernel size of Max--pooling layer. Specifically, we set the kernel size and strides $2\times 2$ for the Max-pooling layer in our experiments as in Table~\ref{tab_set}. As Max-pooling layer chooses the maximum element in $2\times 2$ grids, the other 3 unselected elements do not contribute to the next layer, so the gradients in these locations will be 0 in back propagation. Hence the sparsity of convolutional layer is 25\% at most, and we use 5\% of the full gradients which is also 20\% of 25\%.

\subsection{meProp-CNN with Batch Normalization}
Deep Neural Networks are difficult to train for the reason that the distribution of each layer's inputs changes
during training, as the parameters of the previous layers change. \citet{ioffe2015batch} propose a method called 
Batch Normalization to address this problem. It has been proven to be a very effective technique in Deep Neural 
Networks.

\subsubsection{Batch Normalization}
\citet{ioffe2015batch} find that the distribution of each layer's inputs changes during training, and they refer
 to this phenomenon as \emph{internal covariate shift}. Batch Normalization addresses this problem by performing the
normalization for the training of each mini-batch. \citet{lecun2012efficient, wiesler2011convergence} reveals that the 
network training converges faster if its inputs are whitened – i.e., linearly transformed to have zero
means and unit variances, and decorrelated. By performming normalization to the inputs of each layer, just like
whitening performance, the model would achieve the fixed distributions of inputs that would remove the ill effects
 of the \emph{internal covariate shift}.
Also note that the normalization procedure is different in training and inference. We use mini-batch inputs to compute 
the mean and variance during training, while the unbiased estimatation is used during inference. We use the unbiased variance estimation
$\Var[x] = \frac{m}{m-1}\cdot\E_\setB[\sigma_\setB^2]$, where the expectation is over training mini-batches of size $m$ and
$\sigma_\setB^2$ are their sample variances. 

\subsubsection{Batch-Normalized meProp-CNN}
Batch Normalization achieves great success in Deep Neural Networks, such as deep Convolutional Neural Networks.
Merely adding Batch Normalization to a state-of-the-art image classification model yields a substantial speedup
in training. By further increasing the learning rates, removing Dropout, and applying other modifications afforded
by Batch Normalization, the model reaches the previous state of the art with only a small fraction of training steps~\cite{ioffe2015batch}.
Batch Normalization supplies a new way to regularize the model, just like what Dropout and our proposed method do.
So what if we combine our method \emph{meProp}-CNN with Batch Normalization? Will the Batch-Normalized meProp-CNN could still work properly?
 We test and verify this idea in our experiments and the results are shown as follow.

As decribed in \citet{ioffe2015batch}, Batch Normalization is added before the nonlinearity. For example, 
\begin{equation}
z = g(W\vx+b)
\end{equation}
where $W$ and $\vb$ are learned parameters of the model, and $g(\cdot)$ is the nonlinearity such as sigmoid or
\emph{relu}. The output of $Wx+b$ is normalized before passed to $g(\cdot)$:
\begin{equation}
    z' = g(BN(W\vx+b))
\end{equation}

Batch Normalization method normalizes the outputs of each layer before the activation function. Formally, the process is shown as follows.
We compute the mean and variance over a mini-batch $\setB=\{x_{1\ldots m}\}$:
\begin{align}
\mu_\setB &\leftarrow \frac{1}{m}\sum_{i=1}^m x_i \\
\sigma_\setB^2 &\leftarrow \frac{1}{m}\sum_{i=1}^m (x_i-\mu_\setB)^2
\end{align}
$\mu_\setB$ and $\sigma_\setB^2$ are used to normalize the values of the mini-batch:
\begin{equation}
\xhat_i \leftarrow \frac{x_i-\mu_\setB}{\sqrt{\sigma_\setB^2+\epsilon}}
\end{equation}
$\epsilon$ is a constant added to the mini-batch variance for numerical stability~\cite{ioffe2015batch}. Finally we scale and shift the normalized value:
\begin{equation}
y_i \leftarrow \gamma\xhat_i + \beta
\end{equation}
$\gamma$ and $\beta$ are parameters we should learn during training.

For the convolutional layers, we additionally want the normalization to obey the convolutional property---so that
different elements of the same feature map, at different locations, are normalized in the same way. To achieve this, we jointly normalize all the activations in a minibatch, over all locations.
See Figure~\ref{fig_2} for an illustration, we let $\setB$ be the set of all values in a feature map across both the elements of a mini-batch and spatial locations -- for a mini-batch of size $m$ and feature maps of size $h\times w$, we use the effective mini-batch of size $m'=|\setB| =
m\cdot h\, w$. We learn a pair of parameters $\gamma^\kk$ and $\beta^\kk$ per feature map, rather than per activation.
A side-effect of this constraint is that we should also obey the convolutional property when we apply top-$k$ operation in convolutional layers during back propagation. In other words, we should apply the top-$k$ operation for gradients matrix along with the feature map dimension rather than other dimensions. Take the MNIST experiment as an example: the output of the first convolutional layer is a matrix of size 
$batch\_size\times height\times width \times feature\_map\_num$(that is $m\cdot h\cdot w\cdot c$ in Figure~\ref{fig_2}), during forward propagation we apply batch normalization to each featuremap respectively, which means there are $feature\_map\_num$ pairs of parameters $\gamma^\kk$ and $\beta^\kk$ in our model. Similarly, 
in back propagation we apply the top-$k$ operation for each feature map respectively as the feature maps are computed by filters 
independently.

\begin{table*}[t]
    \centering
      \footnotesize
    \caption{Parameter settings.} \label{tab_set}
    \begin{tabular}{|l|l|l|l|l|l|l|l|}
          \hline
          \multicolumn{2}{|c|}{params} & Conv1 & Pool1 & Conv2 & Pool2 & FC1 & FC2\\
          \hline
          \multirow{2}{*}{MNIST}    &ksize    &$5\times5\times32$  &$2\times2$  &$5\times5\times64$  &$2\times2$  &\multirow{2}{*}{1024}  &\multirow{2}{*}{10}\\
                                    &strides  &$1\times1$          &$2\times2$  &$1\times1$          &$2\times2$  &&\\        
          \hline
    \end{tabular}
    \vspace{-0.05in}
  \end{table*}

\section{Related Work}
\citet{riedmiller1993direct} proposed a direct adaptive method for fast learning, which performs a local adaptation 
of the weight update according to the behavior of the error function. \citet{tollenaere1990supersab} also proposed an adaptive 
acceleration strategy for back propagation. Dropout~\cite{srivastava2014dropout} is proposed to improve training speed and reduce the risk 
of overfitting.
Sparse coding is a class of unsupervised methods for learning sets of over-complete bases to represent data efficiently~\cite{olshausen1996natural}. 
\citet{poultney2007efficient} proposed a sparse autoencoder model for learning sparse over-complete features.
The proposed method is quite different compared with those prior studies on back propagation, dropout, and sparse coding.

The \emph{sampled-output-loss} methods~\cite{jean2014using} are limited to the softmax layer (output layer) and are only based on 
random sampling, while our method does not have those limitations.
The sparsely-gated mixture-of-experts~\cite{shazeer2017outrageously} only sparsifies the mixture-of-experts gated layer and it is limited 
to the specific setting of mixture-of-experts, while our method does not have those limitations.
There are also prior studies focusing on reducing the communication cost in distributed systems~\cite{seide20141,dryden2016communication}, 
by quantizing each value of the gradient from 32-bit float to only 1-bit. Those settings are also different from ours.

\section{Experiments}

To demonstrate the effectiveness of our method, we perform experiments on MNIST image recognition task.
The sample images of the database are shouwn in Figure~\ref{fig_data}. The CNN model without top-$k$ and Batch Normalization is chosen
as baseline.

We implement the proposed method \emph{meProp}-CNN for MNIST~\cite{lecun2010mnist} image recognition task
to verify the method. We use Adam~\cite{kingma2014adam} to optimize the model,
and in our implementation, the detail implementation of Adam is stay untouched. We implement our experiments based on 
Tensorflow~\cite{tensorflow2015-whitepaper}.

\textbf{MNIST:}
The MNIST dataset of handwritten digits has a training set of 60,000 examples, and a test set of 10,000 examples.
The images in this dataset are all gray scale images with size $28 \times 28$ pixel, and they are belong to 10 classes,
which ranges from 0 to 9.


\subsection{Settings}
For MNIST task we use two convolutional layers, two Max--pooling layers and two fully--connected layers, and the output of the
last fully--connected layer is fed to a softmax layer which produces a distribution over the ten-class labels.
The architecture of our model is shown in Figure~\ref{fig_net}.
The first convolutional layer filters the $28\times28$ input image with 32 kernels of size
$5\times5$ with a stride of 1 pixel. The second convolutional layer takes as input
the output of the previous pooling layer and filters it with 64 kernels of size $5\times5$.
The pooling window size of the Max--pooling both are $2\times2$ with a stride of 2 pixels.
We use Rectified Linear Unit (\emph{relu})~\cite{glorot2011deep} as the activation function in our model. \citet{krizhevsky2012imagenet}
find that deep convolutional neural networks with ReLUs converge several times faster than their
equivalents with \emph{tanh} units. Table~\ref{tab_set} shows more details of our parameter setting.

We perform top-$k$ method in back propagation except the last fully--connected layer. The hyper-parameters
of Adam optimization are as follows: the learning rate $\alpha=0.001$, and $\beta_1=0.9, \beta_2=0.999$, $\epsilon=1\times10^{-8}$.
Mini-batch Size is 10.

\subsection{Choice of top-k ratio}
An intuitive idea is that layers with different number of neurons should also have different number of gradients to be selected, and 
the gradients of front layers are influenced by the gradients from the back layers in back propagation, so these factors should be 
taken into account when we set the top-$k$ ratios. In our experiments, We have lots of exploration to get a proper parameters setting.
Experiments reveal that too sparse gradients in back layers(such as fully--connected layers in our experiments) result in bad performance,
which may be that too much gradients information is dropt and this causes the parameters of the front layers can not converge to the 
appropriate values.

\begin{table}[t]
    \centering
    \footnotesize
    \caption{Accuracy results based on \emph{meProp}-CNN of MNIST. Decay means the decay rate we used in momentum method. Epoch means the
    number of epoches to reach the optimal score on development data. The model of this epoch is then used to obtain the test score.
    } \label{tab_meprop}
    \begin{tabular}{|r|l|l l l|}
      \hline
      Top-k(\%)                      &Decay                         &Epoch   &Dev Acc(\%)   &Test Acc(\%)     \\
      \hline
      Baseline                       &$\backslash$                  &18      &99.44         &99.02     \\
      \hline
      \multirow{2}{*}{5\%}           &0                             &28      &99.40         &99.07     \\
                                     &0.6                           &17      &99.40         &99.27     \\
      \hline
      \multirow{2}{*}{8\%}           &0                             &30      &99.42         &99.16      \\
                                     &0.6                           &27      &99.34         &99.23      \\
      \hline
      \multirow{2}{*}{10\%}          &0                             &28      &99.36         &99.15     \\
                                     &0.6                           &19      &99.36         &99.21      \\
      \hline
    \end{tabular}
    \vspace{-0.05in}
  \end{table}

\subsection{Results}
Table~\ref{tab_meprop} shows the results on different top-$k$ values of MNIST dataset. 
The Mini batch size is 10, and the top-$k$ ratio ranges from 5\% to 100\%(the baseline) as shown in the table.
The decay rate represents the tradeoff as in Eq.~\eqref{momentum}. As usual, we first evaluate our model 
on the development data to obtain the optimal number of iterations and the corresponding accuracy, then the test 
data is evaluated based on the best setting tuned in the development set.
As we can see in Table~\ref{tab_meprop}, \emph{meProp}-CNNs get better accuracy than baseline, and the gap of baseline between 
Dev Acc and Test Acc reveals that the baseline without meProp tends to be overfitting. As for momentum method, a higher decay rate does not always mean better result: decay=0.6 works better than 0.9 in our experiments. This may be owing to that too large momentum makes the model inflexible, which means only a small fixed subset of gradients are used while others may never have chances to be selected. Compared with CNN,  \emph{meProp}-CNNs keep the same ability while only keep a small subset of the full gradients in back propagation, or even better.
The main reason could be that the minimal effort update does not modify weakly relevant parameters, which
makes overfitting less likely, similar to the effect of Dropout.

Batch Normalization once again demonstrates the ability to accelerate convergence: the model with Batch Normalization gets a faster rate of convergence and higher accuracy, as shown in Table~\ref{tab_norm}.

\begin{table}[t]
    \centering
    \footnotesize
    \caption{Accuracy results based on \emph{meProp}-CNN with Batch Normalization of MNIST.} \label{tab_norm}
    \begin{tabular}{|r|l|l l l|}
      \hline
      Top-k(\%)                      &Decay                    &Epoch   &Dev Acc(\%)   &Test Acc(\%)  \\
      \hline
      Baseline                       &$\backslash$             &30      &99.60         &99.28     \\
      \hline
      \multirow{2}{*}{5\%}           &0                        &22      &99.58         &99.38     \\
                                     &0.6                      &28      &99.56         &99.39     \\
      \hline
      \multirow{2}{*}{8\%}           &0                        &22      &99.54         &99.26      \\
                                     &0.6                      &22      &99.54         &99.48      \\
      \hline
      \multirow{2}{*}{10\%}          &0                        &23      &99.56         &99.37     \\
                                     &0.6                      &19      &99.52         &99.14      \\
      \hline
    \end{tabular}
    \vspace{-0.05in}
  \end{table}

Batch Normalization with top-$k$ = 5\% gets better accuracy than full gradients. In our experiments the 
gradients of fully--connected layers are not processed by meProp method, and top-$k$ = 5\% means that 5\% gradients are passed back in convolutional layers in back propagation. The momentum \emph{meProp}-CNN with Batch Normalization
is consistent with before, a proper decay rate 0.6 works better in our experiments.
The results are shown in Table~\ref{tab_norm}.

\section{Conclusion and future work}
We propose a new technique called \emph{meProp}-CNN to reduce calculation in back propagation.
In back propagation of CNN, convolution computation is transformed into matrix multiplication operation as in forward propagation, and only a small subset of gradients are used to update the parameters. Specifically,
we select top-$k$ elements to update parameters and the rest are set to $0$, which is similar to the \emph{Dropout} technique. We enhance \emph{meProp} technique with momentum method for more stable results.
Experiments show that our method perform as good as the CNN even only a small subset of gradients are used,
and what's more, it has the ability to avoid overfitting. Our method is still able to work compatibly with Batch Normalization.
In future work, we would like to apply the proposed method to lexical processing tasks~\citep{Gao2010ALS, Sun2008ModelingLI, Sun2010LearningPS, Sun2012FastOT} which may benefit from our method as well.


\bibliography{bib_file}
\bibliographystyle{icml2017}
\end{document}